\documentclass[twoside,11pt]{article}
\pdfoutput=1
\usepackage{jmlr2e}
\usepackage{amsmath,graphicx,amsfonts,amssymb,amscd}
\usepackage{times}
\usepackage{epsfig}
\usepackage{graphicx}
\usepackage{amsmath}
\usepackage{amssymb}
\usepackage{algorithm}
\usepackage{algorithmic}
\usepackage{color,paralist}
\usepackage{comment}
\usepackage{subcaption}
\usepackage{caption}

\newcommand{\R}{\mathbb{R}}

\newcommand{\loss}{\mathcal{L}}
\newcommand{\dist}{\mathcal{D}}

\graphicspath{{figs/}}

\usepackage[pagebackref=true,breaklinks=true,letterpaper=true,colorlinks,bookmarks=false]{hyperref}

\usepackage{lineno}

\newlength\myindent
\begin{document}

\title{Constrained Deep Learning using Conditional Gradient and Applications in Computer Vision
}

\author{\centering Sathya Ravi \quad  Tuan Dinh \quad Vishnu Lokhande \quad Vikas Singh\\
{\tt\small \centering \hspace{25mm} \{sravi, tuandinh, lokhande, vsingh\}@cs.wisc.edu} }
\maketitle
\begin{abstract}
  A number of results have recently demonstrated the benefits
  of incorporating various constraints when training
  deep architectures in vision and machine learning.
  The advantages range from guarantees for statistical
  generalization to better accuracy to compression.
  But support for general constraints within widely
  used libraries remains scarce and their broader
  deployment within many applications that can benefit
  from them remains under-explored. Part of the reason is
  that Stochastic gradient descent (SGD), the workhorse for
  training deep neural networks, does not natively deal with constraints
  with global scope very well. In this paper, we revisit a classical first order
  scheme from numerical optimization, Conditional Gradients (CG), that has, thus far had limited applicability
  in training deep models. We show via rigorous
  analysis how various constraints can be naturally handled by  modifications 
  of this algorithm. We provide convergence guarantees and show a suite of
  immediate benefits that are possible --- from training ResNets with fewer layers but better
  accuracy simply by substituting in  our version of CG to faster training of GANs with 50\% fewer
  epochs in image inpainting applications to provably better generalization guarantees using
  efficiently implementable forms of recently proposed regularizers. 

\end{abstract}


\section{Introduction}
The learning or fitting problem in deep neural networks in the supervised setting
is often expressed as the following stochastic optimization problem,\begin{align}
\min_W\mathop{\mathbb{E}} \limits_{(x,y)\sim\dist}\loss(W;(x,y))\label{unreg_dl}
\end{align}
where $W=W_1\times \cdots \times W_l$ denotes the Cartesian product of the weight matrices of the
network with $l$ layers that we seek to learn from the data $(x,y)$ sampled from the underlying distribution $\dist$.
Here, $x$ can be thought of the ``features'' (or predictor variables) of the data and $y$ denotes the ``labels'' (or the response variable).
The variable $W$ parameterizes the function that we desire to learn that predicts the labels given
the features whose accuracy is measured using the loss function $\loss$.
In the unsupervised  setting, we do not have a notion of labels, hence the common practice
is to inject some prior knowledge or inductive bias into the loss $\loss$. 
For simplicity, the specification above is intentionally transparent to the activation
function we use between the layers and the specific network architecture.
Most common instantiations of the above task are non-convex but results
in the last five years demonstrate that reasonable minimizers that
generalize well to test data can be found via SGD and
its variants. Recent results have also explored the interplay between
the overparameterization of the network, its degrees of freedom and issues related to global optimality \cite{haeffele2015global,janzamin2015beating,soudry2016no}. 

{\bf Regularizers.} Independent of the network architecture we choose to deploy for a given task,
a practitioner may often want to impose additional constraints or regularizers, pertinent to
the application domain of interest. 
In fact, the use of task specific constraints to improve the behavioral performance of \emph{neural networks}, both
from a computational and statistical perspective, has a long history dating back at least to the 1980s \cite{platt1988constrained,zhang1992lagrange}.
These ideas are being revisited recently \cite{rudd2014constrained} motivated by generalization, convergence or simply as a strategy for compression \cite{cheng2017survey,han2015deep}. 
However, using constraints on the types of architectures that are common in modern computer vision problems
is still being actively researched by various groups.
For example, \cite{mikolov2014learning} demonstrated that training Recurrent Networks can be accelerated by constraining
a part of the recurrent matrix to be close to identity. Sparsity and low-rank encouraging constraints have shown promise in
a number of settings \cite{tai2015convolutional,liu2015sparse}. In an interesting paper,
\cite{pathak2015constrained} showed that linear constraints on the output layer improves the accuracy on a
semantic image segmentation task. \cite{marquez2017imposing} showed that hard constraints on the output layer yield competitive results on the 3D human pose estimation task and \cite{oktay2017anatomically} used anatomical constraints for cardiac image analysis.
The above discussion suggests that while there are some results demonstrating the value of {\em specific} constraints for
{\em specific} problems, the development is still in a nascent stage. It is, therefore, not surprising that 
the existing software libraries and APIs for deep learning (DL) in vision and machine learning offer little to no support
for constraints. For example, 
Keras only offers support for simple bound constraints\cite{Charles2013}.

{\bf Optimization Schemes.} Let us set aside the issue of constraints for a moment and discuss the choice of the optimization schemes that are
in use today. There is little doubt that SGD algorithms dominate
the landscape of DL problems in vision and machine learning.
Instead of evaluating the loss and the gradient over the full training set,
SGD simply computes the gradient of the parameters using a few training examples. It mitigates
the cost of running back propagation over the full training data and comes with various theoretical
guarantees as well \cite{hardt2016train,raginsky2017non}. The reader will notice that part of the reason
that constraints have  not been intensively explored in a broader range of problems
may have to do with the interplay between constraints and the SGD algorithm \cite{marquez2017imposing}. 
While some regularizers and ``local'' constraints are easily handled within SGD, some others require
a great deal of care and can adversely affect convergence and practical runtime \cite{bengio2012practical}. There are
also a broad range of constraints where SGD is unlikely to work well based on theoretical results known today --- and it remains an
open question in optimization \cite{johnson2013accelerating,defazio2014saga}. 
We note that algorithms other than the standard SGD have remained a constant focus of research in the community
since they offer many theoretical advantages that can also be easily translated to practice \cite{dauphin2015equilibrated,grubb2010boosted}.
These include adaptive sub-gradient methods such as Adagrad \cite{duchi2011adaptive}, the RMSprop algorithm \cite{dauphin2015equilibrated}
which addresses the issue of ill-conditioning in Deep Networks with a normalized form of SGD, and
various adaptive schemes for learning rate adjustments \cite{zeiler2012adadelta} and utilizing the momentum
method \cite{kingma2014adam}. However, the reader will notice that these methods only impose constraints in a ``local'' fashion since the computational cost of imposing global constraints using SGD-based methods becomes extremely high \cite{pathak2015constrained}. We explore this issue in depth in the next section.

\section{Why do we need to impose constraints?}
The question of why constraints are needed for statistical learning models in vision and machine learning can be
equivalently restated in terms of the need for regularization in setting up learning models --- a notion that is well known
but will be precisely stated shortly. Recall that
regularization schemes in one form or another
go nearly as far back as the study of fitting models to observations of data \cite{wahba1990spline,mundle1959probability}.
Broadly speaking, such schemes can be divided into two related categories: algebraic and statistical (or probabilistic). 
The first category may refer to problems that are otherwise not possible or difficult to solve, also known as ill-posed problems \cite{tikhonov1987ill}.
For example, without introducing some additional piece of information, it is not possible to solve a linear system of equations $Ax=b$ in which
the number of observations (rows of $A$) is less than the number of degrees of freedom (columns of $A$). 
In the second category, one may use regularization as a way of ``explaining'' data using simple hypotheses rather than complex ones, for example,
the minimum description length principle \cite{rissanen1985minimum}. The underlying rationale is that, complex hypotheses are less likely to be accurate
on the unobserved samples (also known as Occam's razor). In other words, the generalizability of models diminish as the complexity of the model
increases for a fixed dataset size. Or equivalently, we need more data to train complex models. Recent developments on the theoretical side of DL showed that imposing simple but \emph{global} constraints on the parameter space of deep networks is an effective way  of analyzing the learning theoretic properties including sample complexity and generalization error \cite{neyshabur2015norm}. 
Hence we seek to solve, \begin{align}
\min_W\mathop{\mathbb{E}} \limits_{(x,y)\sim\dist}\loss(W;(x,y)) + \mu R(W)\label{reg_dl}
\end{align}
where $R(\cdot)$ is a suitable regularization function for a fixed $\mu>0$.
We usually assume that $R(\cdot)$ is simple, for example, in the sense that the gradient can be computed efficiently. Using the Lagrangian interpretation,
Problem \eqref{reg_dl} is the same as the following constrained formulation,\begin{align}
\min_W\mathop{\mathbb{E}} \limits_{(x,y)\sim\dist}\loss(W;(x,y))~\text{s.t.}~ R(W)\leq \lambda \label{cons_dl}
\end{align} 
where $\lambda>0$. Note that when the loss function $\mathcal{L}$ is convex, both the above problems  are equivalent in the sense that given $\mu>0$ in \eqref{reg_dl}, there exists a $\lambda>0$ in \eqref{cons_dl} such that
the {\em optimal solutions to both the problems coincide} (see sec 1.2 in \cite{bach2012optimization}). In practice, both $\lambda$ and $\mu$ are chosen by standard statistical procedures such as cross validation. 

{\bf Finding Pareto Optimal Solutions:} On the other hand, when the loss function is nonconvex as is typically the case in DL, formulation \eqref{cons_dl} is more powerful than \eqref{reg_dl}. Let us see why: for a fixed $\lambda>0$, there might be solutions $W^*_\lambda$ of \eqref{cons_dl} for which there exists \emph{no} $\mu>0$ such that $W^*_{\lambda}=W^*_{\mu}$ whereas any solution of problem \eqref{reg_dl} can be obtained for some $\mu$ in \eqref{cons_dl}, see section 4.7 in \cite{boyd2004convex}. It turns out that it is easier to understand this phenomenon through the lens of multiobjective optimization.  In multiobjective optimization, care has to be taken to even define the notion of optimality of feasible points (let alone computing them efficiently) depending on the problem. Among various notions of optimality, we will now argue that Pareto optimality is the most suited for our goal.  

Recall that our goal is to  find $W$'s that achieve low training error and are at the same time ``simple'' (as measured by $R$).  In this context, a pareto optimal solution is a point $W$ such that none of $\mathcal{L}(W)$ or $R(W)$ in \eqref{cons_dl} or \eqref{reg_dl} can be made better without making the other worse, thus capturing the essence of overfitting effectively.  In practice, there are many algorithms to find pareto optimal solutions and this is where problem \eqref{cons_dl} dominates \eqref{reg_dl}. Specifically, formulation \eqref{reg_dl} falls under the category of ``scalarization'' technique whereas \eqref{cons_dl} is $\epsilon$-constrained technique. It is well known that when the problem is nonconvex, $\epsilon$-constrained technique yields pareto optimal solutions whereas scalarization technique does not!

Finally, we should note that even when the problems \eqref{reg_dl} and \eqref{cons_dl} are equivalent, in practice, algorithms that are used to solve them can be very different.\\
{\bf Our contributions:} We show that many interesting global constraints of interest in vision/machine learning after reformulations can be enforced using a classical optimization technique that has not been deployed  much at all in training deep learning models. We analyze the theoretical aspects of this proposal in detail. On the application side, specifically, we explore and analyze the performance of our Conditional Gradient (CG) algorithm with a specific focus on training deep models on the constrained formulation shown in \eqref{cons_dl}. Progressively, 
we go from cases where there is no (or negligible) loss of both accuracy and training time to scenarios where this procedure shines
and offers significant benefits in performance, runtime and generalization. Our experiments indicate that: (i) {\bf with less than $50\%$ \#-parameters}, we improve ResNet accuracy by $25\%$ (from $8\% $ to $6\%$ test error), and (ii) GANs can be trained in nearly a {\bf third of the computational time } achieving the same or better qualitative performance on an image inpainting task.  
\section{First Order Methods: The Two Towers} To setup the stage for our development we first discuss the two broad strategies that are used to solve problems of the form shown in \eqref{cons_dl}.
{\bf First}, a natural extension of gradient descent (GD) also known as Projected GD (PGD)   may be used (over multiple iterations). Intuitively, we take a gradient step and then compute the point
that is closest to the feasible set defined by the regularization function.  Hence, at each iteration PGD requires the solution of the following optimization problem or the so-called Projection operator,\begin{align}
W_{t+1}^{PGD}\leftarrow \arg\min_{W: R(W)\leq \lambda} \frac{1}{2} \|W-\left(W_t-\eta g_t \right)\|_F^2 \label{pgd_1}
\end{align}
where $\|W\|_F$ is the Frobenius norm of $W$, $g_t$ is (an estimate of) the gradient of $\loss$ at $W_t$ and $\eta$ is the step size or learning rate. In practice, we compute $g_t$ by using only a few training samples (or minibatch) and running the backpropagation algorithm. Note that the objective function  $\mathbb{E}_{(x,y)\sim \mathcal{D}}\mathcal{L}(W)$ is smooth in $W$ for any probability distribution $\mathcal{D}$ and is commonly referred to as \emph{stochastic smoothing}. Hence, for our descriptions, we will assume that the derivative is well defined \cite{ghadimi2013stochastic}.
Furthermore, when there are no constraints, \eqref{pgd_1} is simply the standard SGD method broadly used in the literature.
Essentially, we can see that \eqref{pgd_1}  requires optimizing a quadratic function on the feasible set. So,
the main bottleneck while imposing constraints with PGD is the complexity of solving \eqref{pgd_1}.
Even though many $R(\cdot)$ do admit an efficient procedure in theory, using them for applications in training deep models has been a challenge
since they may be complicated or not easily amenable to a GPU implementation \cite{taylor2016training,frerix2017proximal,wongneural}. 

So, a natural question to ask is whether there are methods that are faster in the following sense: can we solve simpler problems at each iteration and
also impose the constraints effectively? An assertive answer is provided by a scheme that falls under the {\bf second}
general category of first order methods: the Conditional Gradient (CG) algorithm \cite{reddi2016stochastic,frank1956algorithm}.
Recall that CG methods solve the following {\em linear minimization problem at each iteration} instead of a quadratic one\begin{align}
s_t \in \arg \min_{W} g_t^TW ~\text{s.t.}~R(W)\leq \lambda\label{main_cg}
\end{align}
and update $W_{t+1}^{CG} \leftarrow \eta W_t + (1-\eta)s_t$.  While both PGD and CG guarantee convergence with mild conditions on $\eta$,
it may be the case (as we will see shortly) that problems of the form \eqref{main_cg} can be {\em much simpler} than the form in \eqref{pgd_1} and hence
suitable for training deep learning models. An additional bonus is that CG algorithms also offer nice space complexity guarantees that are also tangibly attainable
in practice, making it a very promising choice for \emph{constrained} training in deep models. 

\begin{remark}\label{cg_compact}
	Note that in order for CG algorithm \eqref{main_cg} to be well defined, we need the feasible set to be bounded whereas this is not required for PGD \eqref{pgd_1}. 
\end{remark}
In order to study the  promise of the algorithm for training deep models,
it is important to specifically understand exactly how the CG algorithm behaves for the regularization constraints that are commonly used in vision and machine learning.
\begin{remark}
Note that although the loss function $\mathbb{E}\mathcal{L}(W)$ is nonconvex, the constraints that we need for almost all applications are convex, hence, all our algorithms are guaranteed to converge by design \cite{lacoste2016convergence,reddi2016stochastic} to a stationary point.
\end{remark}
\section{Categorizing ``Generic'' Constraints for CG}
In this section, we describe how a broad basket of ``generic'' constraints 
broadly used in our community, can be arranged into a hierarchy of sorts --- ranging from cases where a CG scheme is perfect and expected to yield wide-ranging
improvements to situations where the performance is only satisfactory and additional technical development is warranted. 
For example, the $\ell_1$-norm is often used to induce sparsity \cite{collins2014memory}. The nuclear norm (sum of singular values) is used to induce a low rank regularization, often
for compression and/or speed-up reasons \cite{barone2016low}. 

So, how do we know which constraints when imposed using CG are likely to work well? In order to analyze the qualitative nature of constraints suitable for CG algorithm,
we categorize the constraints into three main categories based on how the update (in each iteration) will computationally, and learning-wise compare to a
SGD update \emph{with} these constraints.
\subsection{Category 1 constraints are excellent}
We categorize constraints as Category 1 
if both the SGD and CG updates take a similar form algebraically. The reason we call this category ``excellent'' is because it is easy
to transfer the empirical knowledge that we obtained in the unconstrained setting, specifically, learning and dropout rates to the regime where
we want to impose these additional constraints.
In this case, we see that we get quantifiable improvements in terms of both computation and learning.

Two types of generic constraints fall into this category: {\bf 1)} the Frobenius norm and {\bf 2)} the Nuclear norm \cite{ruder2017overview}.
We will now see how we can solve \eqref{pgd_1} and \eqref{main_cg} by comparing and contrasting them. 

{\bf Frobenius Norm.} When $R(\cdot)$ is the Frobenius norm, it is easy to see that \eqref{pgd_1} corresponds to the following,\begin{align}
W_{t+1}^{PGD} = \begin{cases}
W_t- \eta g_t & \text{if} ~~ \|W_t- \eta g_t\|_F\leq \lambda \\
\lambda\cdot\frac{W_t- \eta g_t}{\|W_t- \eta g_t\|_F} & \text{otherwise},
\end{cases}
\end{align}
and \eqref{main_cg} corresponds to $s_t = -\lambda\frac{g_t}{\|g_t\|_F}$ which implies that,\begin{align}
W_{t+1}^{CG}=  W_t - (1-\eta)\left(W_t+\lambda\frac{g_t}{\|g_t\|_F}\right)\label{cg_fro}.
\end{align}
It is easy to see that both the update rules essentially take the same amount of calculation which can be easily done while performing a backpropagation step.
So, the actual change in any existing implementation will be minimal but CG will automatically offer an important advantage, notably {\bf scale invariance}, which
several recent papers have found to be advantageous -- both computationally and theoretically \cite{lacoste2015global}. 

{\bf Nuclear norm.} On the other hand, when $R(\cdot)$ is the nuclear norm, the situation where we use CG (versus not) is quite different.
All known projection (or proximal) algorithms require computing at each iteration the full singular value decomposition of $W$, which in the case
of deep learning methods becomes restrictive \cite{recht2010guaranteed,cai2010singular}. In contrast, CG only requires computing the top-$1$ singular vector of $W$
which can be done easily and efficiently on a GPU via the power method \cite{jaggi2013revisiting}. Hence in this case, if the number of edges in
  the network is $|E|$ we get a {\em near-quadratic speed up}, i.e., from $O(|E|^3)$ for PGD to $O(|E|\log |E|)$ making it practically implementable \cite{golub2012matrix} in
the very large scale settings encountered in vision. 
Furthermore, it is interesting to observe that the rank of $W$ after running $T$ iterations of CG is at
most $T$ which implies that we need to only store $2T$ vectors instead of the whole matrix $W$ making it a viable solution for deployment on small form factor devices with
memory constraints \cite{howard2017mobilenets}.
Hence, in this case we can obtain a strong practical impact of CG algorithms immediately. The main takeaway is that, since projections are computationally expensive, projected SGD  is not a viable option in practice.

\subsection{Category 2 constraints are potentially good} \label{weak_acpt}
As we saw earlier, CG algorithms are always {\em at least} as efficient as the PGD updates:
in general, any constraint that can be imposed using the PGD algorithm can also be imposed by CG algorithm, if not faster.
Hence, generic constraints are defined to be Category 2 constraints for CG if the empirical knowledge cannot be
easily transferred from PGD. Two classical norms that fall into this category: $\|W\|_1$ and $\|W\|_{\infty}$.
For example, PGD on the $\ell_1$ ball can be done in linear time (see \cite{duchi2008efficient}) and for $\|W\|_{\infty}$ using gradient clipping \cite{boyd2004convex}.
So, let us evaluate the CG step \eqref{main_cg} for the constraint $\|W\|_1\leq \lambda$ which corresponds to, \begin{align}
s_t^j=\begin{cases}\label{cg_l1}
-\lambda & \text{if} ~~ j^*=\arg\max\limits_{j}\left|g_t^j\right| \\
0 & \text{otherwise}.
\end{cases}
\end{align}
That is, we assign $-\lambda$ to the coordinate of the gradient $g_t$ that has the maximum magnitude in the gradient matrix.
We see that this exactly corresponds to a \emph{deterministic} dropout regularization in which at each iteration we only update {\em one} edge of the network. 
While this might not be necessarily bad, it is now common knowledge that a high dropout rate (i.e., updating very few weights at each iteration) leads to underfitting or in other words,
the network tends to need a longer training time \cite{srivastava2014dropout}.  Similarly, the update step \eqref{main_cg} for CG algorithm with $\|W\|_{\infty}$ takes the following form,\begin{align}
s_t^j=\begin{cases}\label{cg_linf}
+\lambda & \text{if} ~~ g_t^j < 0  \\
-\lambda & \text{otherwise}.
\end{cases}
\end{align}
In this case, the CG update uses only the sign of the gradient and does not use the magnitude at all.
In both cases, one issue is that information about the gradients is not used by the standard form of the
algorithm making it not so efficient for practical purposes. 
Interestingly, even though the update rules in \eqref{cg_l1} and \eqref{cg_linf} use extreme ways of using the gradient information,
we can, in fact, use a group norm type penalty to model the trade-off. Recent work shows that there are very efficient procedures to solve the corresponding CG updates as well \eqref{main_cg}. For space reasons, these ideas are described in the supplement and \cite{garber2015faster}.
\begin{remark}
The main takeaway from the discussion is that Category 2 constraints surprisingly unifies many regularization techniques that are traditionally used in DL in a more methodical way.
\end{remark}

\subsection{Category 3 constraints need more work}
There is one class of regularization norms that do not nicely fall in either of the above categories,
but is used in several problems in vision: the Total Variation (TV) norm. TV norm is widely used in denoising algorithms
to promote smoothness of the estimated sharp image \cite{chambolle1997image}.
The TV norm on an image $I$ is defined as a certain type of norm of its discrete gradient field $(\nabla_iI(\cdot),\nabla_jI(\cdot))$ i.e.,
\begin{align}
\|I\|_{TV}^{p}:=(\left(\|\nabla_iI\|_p+\|\nabla_jI\|_p\right)^{p}.
\end{align}
Note that for $p \in \{1,2\}$, this corresponds to the classical anisotropic and isotropic TV norm respectively.
Motivated by the above idea, we can now define the TV norm of a Feed Forward Deep Network. TV norm, as the name suggests, captures the notion of {balanced networks}, shown to make the network more stable \cite{neyshabur2015path}.
Let $A$ be the incidence matrix of the network: the rows of $A$ are indexed by the nodes and the columns are indexed by the (directed) edges such
that each column contains exactly two nonzero entries: a $+1,-1$ in the rows corresponding to the starting node $u$ and ending node $v$ respectively.
Let us also consider the weight matrix of the network  as a vector (for simplicity) indexed in the same order as the columns of $A$.
Then, the TV norm of the deep neural network is, \begin{align}
\|W\|_{TV}:=\|AW\|_p\label{deep_tv}.
\end{align}
It turns out that when $R(W)=\|W\|_{TV}$, PGD is {\bf not} trivial to solve and requires special schemes \cite{fadili2011total}
with runtime complexity of $O(n^4)$ where $n$ is the number of nodes --- impractical for most deep learning applications in vision.
In contrast, CG iterations only require a special form of maximum flow computation which can be done efficiently \cite{goldfarb2009parametric,harchaoui2015conditional}. 
\begin{lemma}
An $\epsilon$-approximate CG step \eqref{main_cg} can be computed  in $O(1/\epsilon)$ time (independent of dimensions of $A$).
\begin{proof} \emph{(Sketch)} We show that the problem is equivalent to solving the dual of a specific linear program. This can be efficiently accomplished using \cite{johnson2013accelerating}. Due to space reasons, the full proof is in the supplement.
\end{proof}
\end{lemma}
\begin{remark}
  The above discussion suggests that conceptually, Category 3 constraints can be incorporated
  and will immensely benefit from CG methods. However,
  unlike Category 1-2 constraints, it requires specialized implementations to solve subproblems  from \eqref{deep_tv}  which
  are not currently available in popular libraries. So, additional work is needed before broad utilization may be possible. 
\end{remark}

\section{Path Norm Constraints in Deep Learning}
So far, we only covered constraints that were already in use in vision/machine learning and recently, some attempts \cite{marquez2017imposing} were made to utilize them in 
deep networks. 
Now, we review a new notion of regularization, introduced very recently, that has its roots primarily in deep learning \cite{neyshabur2015path}.
We will first see the definition and explain some of the key properties that this type of constraint captures. 
\begin{definition}
\cite{neyshabur2015path} The $\ell_2$-path regularizer is defined as :\begin{align}
\|W\|_{\pi}^2 =\sum_{v_{in}[i]\xrightarrow{e_1}v_1\xrightarrow{e_2}\cdots v_{out}[j]} \left|\prod_{j=1}^l W_{e_k}  \right|^2.\label{path_norm_def}
\end{align}
\end{definition}
Here $\pi$ denotes the set of paths, $v_{in}$ corresponds to a node in the input layer, $e_i$ corresponds to an edge between a node $(i-1)$-th layer and $i$-th layer that lies in
the path between $v_{in}$ and $v_{out}$ in the output layer. Therefore, the path norm measures {\em norm of all possible paths $\pi$ in the network up to the output layer}.
\begin{algorithm}[!t]
	\caption{ \label{alg:path_cg} Path-CG iterations}
	\begin{algorithmic}
		\STATE Pick a starting point  $W_0:\|W_0\|_{\pi}\leq \lambda$ and $\eta\in (0,1)$.
		\FOR{$t = 0,1,2,\cdots,T$ iterations}		
		\FOR{$j = 0,1,2,\cdots,l$ layers}		
		\STATE  $g\leftarrow$ gradient of edges from $j-1$ to $j$ layer.
		\STATE Compute $\gamma_e~\forall~ e$ from $j-1$ to $j$ layer (eq \eqref{path_gamma_e})
		\STATE Set $s^j_t \leftarrow \arg\min_{W} g^TW\text{s.t.}\|\Gamma W\|_2\leq \lambda$ (eq\eqref{cg_path})
		\STATE Update $W_{t+1}^j\leftarrow \eta W_t^j + (1-\eta )s_t^j$
		\ENDFOR				
		\ENDFOR
	\end{algorithmic}
\end{algorithm}

{\bf Why do we need path norm?} One of the basic properties of ReLu (Rectified Linear Units)  is that it is \emph{scaling invariant} in the following way: multiplying the weights of incoming edges to a node $i$ by a positive constant and dividing the outgoing edges from the same node $i$ does not change $\loss$ for any $(x,y)$. Hence,
an update scheme that is {\em scaling invariant} will significantly increase the training speed. Furthermore, the authors in \cite{neyshabur2015path}
showed how path regularization converges to optimal solutions that can generalize better compared to the usual SGD updates --- so apart from
computational benefits, there are clear statistical generalization advantages too. 

{\bf How do we incorporate the path norm constraint?}  Recall from Remark \ref{cg_compact} that the feasible set has to be bounded, so that the step \eqref{main_cg} is well defined.
Unfortunately, this is not the case with the path norm. To see this, consider a simple line graph with weights $W_1$ and $W_2$.
In this case, there is only one path and the path norm constraint is $W_1^2W_2^2 \leq 1$ which is clearly unbounded.
Further, we are not aware of an efficient procedure to compute the projection for higher dimensions
since there is no known efficient separation oracle. Interestingly, we take advantage of the fact that if we fix $W_1$, then the feasible set is bounded.
This intuition can be generalized, that is, we can update one layer at a time which we will describe now precisely.
\begin{figure*}[!t]
\centering
\includegraphics[width=0.95\linewidth]{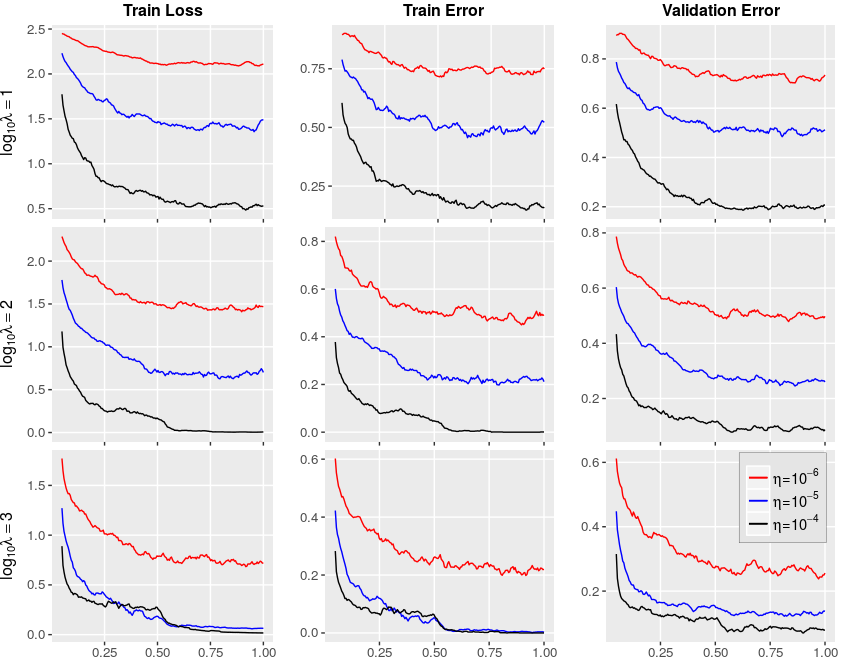}
\vspace{-12pt}
\caption{\label{fig:resnet1}\footnotesize Performance of CG on ResNet-32 on CIFAR10 dataset ($x$-axis denotes the fraction of $T$): as $\lambda$ increases, the training error, loss value and test error all start to decrease simultaneously. Note that if $\lambda$ is sufficiently large (not over constraining the network), training and test errors start to go down significantly faster for a wide range of stepsizes $\eta$, showing that CG is stable under $\eta$.}
\vspace{-12pt}
\end{figure*}

{\bf Path-CG Algorithm:} In order to simplify the presentation, we will assume that there are no biases noting that
the procedure can be easily extended to the case when we have individual bias for every node.
Let us fix a layer $j$ and the vectorized weight matrix of that layer be ${W}$ that we want to update and as usual,
${g}$ corresponds to the gradient. Let the number of nodes in the $(j-1)$ and $j$-th layers be $n_1$ and $n_2$ respectively.
For each edge between these two layers we will compute the scaling factors $\gamma_e$  defined as,\begin{align}
\gamma_e = \sum_{v_{in}[i]\cdots\xrightarrow{e}\cdots v_{out}[j]} \left|\prod_{e_k\neq e} W_{e_k}  \right|^2.\label{path_gamma_e}
\end{align} Intuitively, $\gamma_e$ computes the norm of all paths that pass through the edge $e$ excluding the weight of $e$.
This can be efficiently done using Dynamic Programming in time $O(l)$ where $l$ is the number of layers. Consequently, the computation of path norm also satisfies the same runtime, see \cite{neyshabur2015path} for more details.
Now, observe that the path norm constraint when all of the other layers are
fixed reduces to solving the following problem (the detailed derivation is in the supplement),\begin{align}
\min_{W} g^TW~\text{s.t.}~\|\Gamma W\|_2\leq \lambda\label{cg_path}
\end{align}
where $\Gamma$ is a diagonal matrix with $\Gamma_{e,e}=\gamma_e$, see \eqref{path_gamma_e}.
Hence, we can see that the problem again reduces to a simple rescaling and then normalization as seen for the Frobenius norm in \eqref{cg_fro} and repeat for each layer.
\begin{remark}
  The starting point $W_0$ such that $\|W_0\|_{\pi}\leq \lambda$ can be chosen simply by randomly assigning the weights from the Normal Distribution with mean $0$. 
\end{remark}

{\bf Complexity of Path-CG \ref{alg:path_cg}:} From the above discussion, our full algorithm is given in Algorithm \ref{alg:path_cg}.
The main computational complexity in Path-CG comes from computing the matrix $\Gamma$ for each layer, but as we described
earlier, this can be done {\em by} backpropagation. Hence, the complexity of our algorithm for running $T$ iterations   is essentially $O(lBT)$ where $B$ is a size of the mini-batch.

{\bf Scale invariance of Path-CG \ref{alg:path_cg}:} Note that CG algorithms satisfy  a much general property called as Affine Invariance \cite{jaggi2013revisiting}, which implies that
it is also scale invariant. Scale invariance makes our algorithm more efficient (in wall clock time) since it avoids exploring functions that compute the value.

\section{Experimental Evaluation}
\begin{figure*}[!t]
	\begin{center}
		\includegraphics[width=0.95\linewidth]{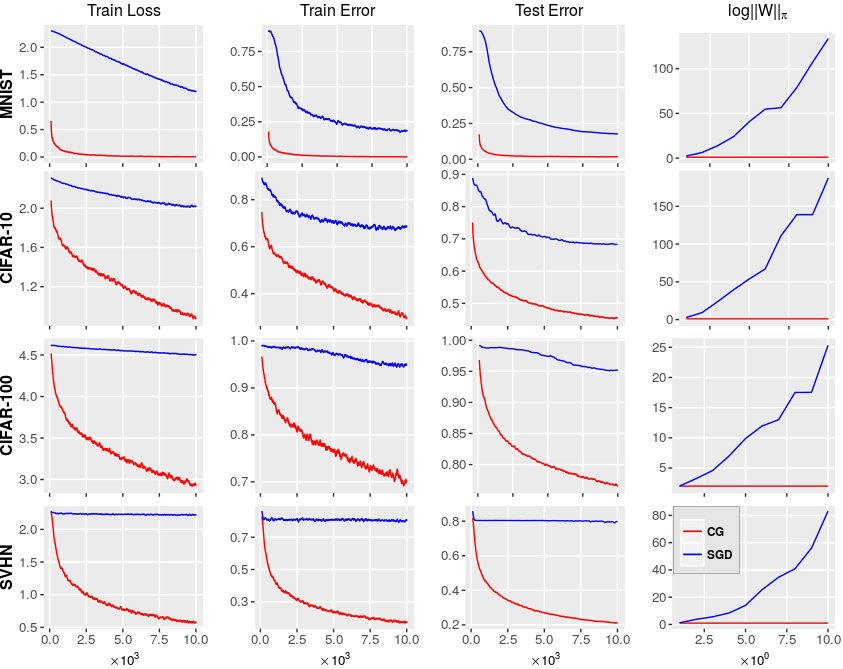}
	\end{center}
        \vspace{-14pt}
	\caption{\label{fig:pathcg}\footnotesize Performance of Path-CG vs SGD on a 2-layer fully connected network on four datasets (x-axis denotes the $\#$iterations). Observe that across all datasets, Path-CG is much faster than SGD (first three columns). Last column shows that SGD is not stable with respect to the path norm.}
        \vspace{-12pt}
\end{figure*}
We present experimental results on three different case studies to support our basic premise and theoretical findings in the earlier sections: constraints can be easily handled with our CG algorithm in the context of Deep Learning while preserving the empirical performance of the models.
{\bf The first set} of experiments is designed to  show how simple/generic constraints can be easily incorporated in existing deep learning models to get both faster training times and better accuracy while reducing the \#-layers using the ResNet architecture.
{\bf The second set} of experiments is to evaluate our Path-CG algorithm. The goal is to show that Path-CG is much more stable than the Path-SGD algorithm in \cite{neyshabur2015path}, implying lower generalization error of the model. 
{\bf In the third set} of experiments we show that GANs (Generative Adversarial Networks) can be trained faster using the CG algorithm and that the training tends to be stable. To validate this, we test the performance of the GAN  on an image inpainting application. Since CG algorithm maintains a solution that is a  convex combination of all previous iterates, hence to decrease the effect of random initialization,  the training scheme consists of two phases: (i) burn-in phase in which the CG algorithm is run with a constant stepsize; (ii) decay phase in which the stepsize is decaying according to $1/t$. This makes sure that the effect of randomness from the initialization is diminished. We use $1$ epoch for the burn-in phase, hence we can conclude that the algorithm is guaranteed to converge to a stationary point\cite{lacoste2016convergence}.


\subsection{Improve ResNets using Conditional Gradients}
We start with the problem of  image classification, detection and localization. For these tasks, one of the best performing architectures are variants of the Deep Residual Networks (ResNet) \cite{he2016deep}. For our purposes, to analyze the performance of CG algorithm, we used the shallower  variant of ResNet, namely ResNet-32 (32 hidden layers) architecture and trained on the CIFAR10 \cite{krizhevsky2012imagenet} dataset.  ResNet-32 consists of $5$ residual blocks and $2$ fully connected, one each at the input and output layers. Each residual block consists of 2 convolution, ReLu (Rectified Linear units), and batch normalization layers, see \cite{he2016deep} for more details. 
CIFAR10 dataset contains  $60000$ color images of size $32\times 32$ with $10$ different categories/labels. Hence, the network contains approximately $0.46M$ parameters. 

To make the discussion clear, we present results for the case where the total Frobenius norm of the network parameters is constrained to be less than $\lambda$ and trained using the CG algorithm. To see the effect of the parameters $\lambda$ and step sizes $\eta$ on the model, we ran $80000$ iterations, see Figure  \ref{fig:resnet1}. The plots essentially show that if $\lambda$ is chosen big enough, then the accuracy of CG is very close to the accuracy of ResNet-164 ($5.46\%$ top-1 test error, see \cite{he2016deep}) that has many more parameters (approximately 5 times!). This is an interesting property with immediate practical import and shows that CG can be used to improve the performance of \emph{existing architectures} by appropriately choosing constraints (see supplement for more experiments).

\emph{{\bf Takeaway:} CG offers fewer parameters and higher accuracy on a standard network with no additional change.}

\begin{figure*}[!t]
	\centering
	\begin{subfigure}[!t]{0.47\textwidth}
		\includegraphics[width=1\textwidth]{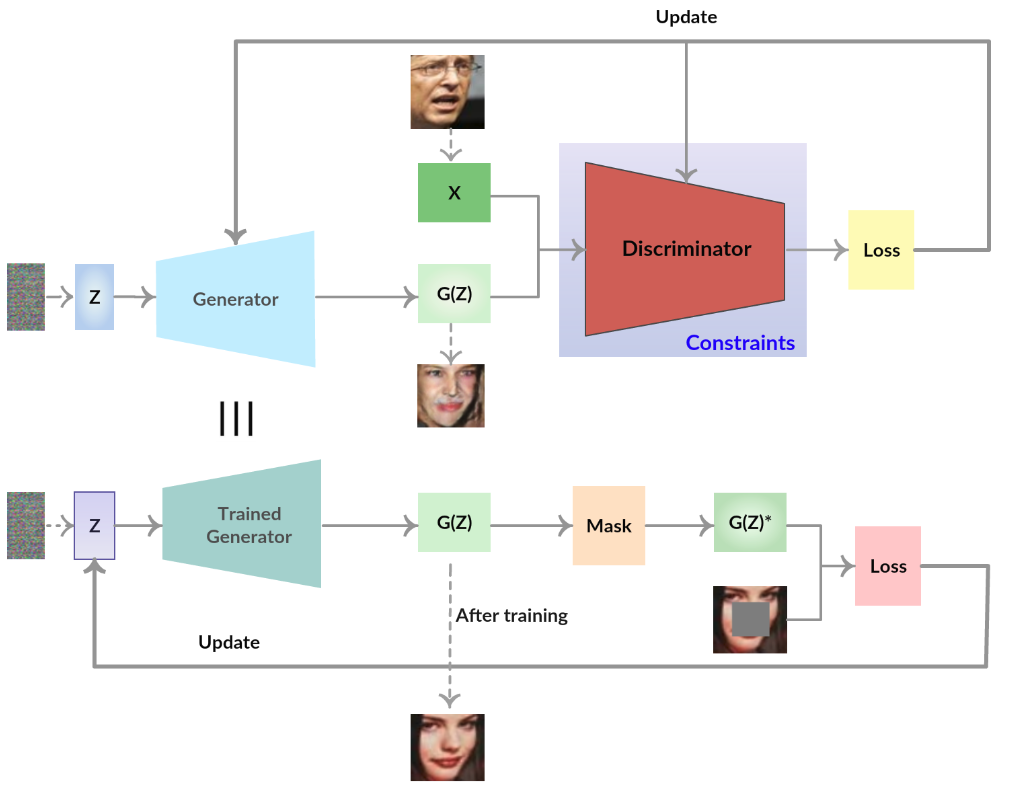}
	\end{subfigure}
	\begin{subfigure}[!t]{0.47\textwidth}
		\includegraphics[width=1\textwidth]{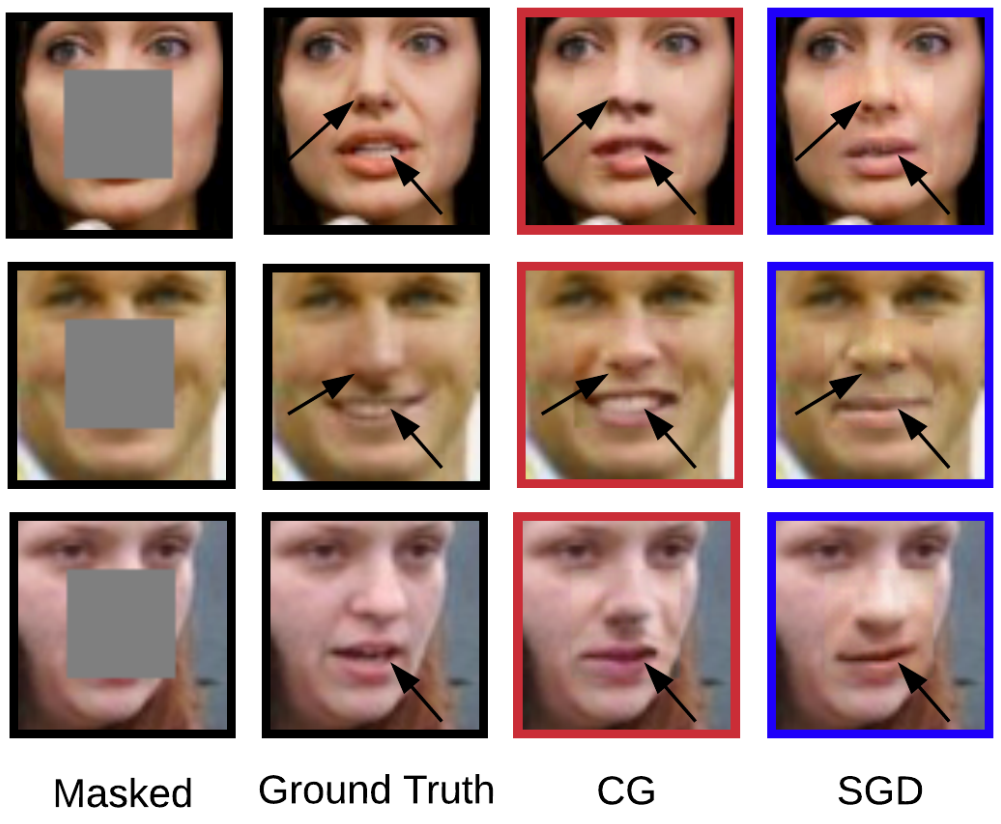}
	\end{subfigure}
        \vspace{-10pt}
	\caption{	\label{fig:inpaint_ill}{\bf Left:} \footnotesize Illustrates the task of image inpainting overall pipeline. {\bf Right:} CG-trained DC-GAN performs as good as (or better than) SGD-based DC-GAN but with $50\%$ epochs (more qualitative results are provided in the supplement).}
        \vspace{-9pt}
\end{figure*}
\subsection{Path-CG vs Path-SGD: Which is better?}
In this case study, the goal is to compare Path-CG with the Path-SGD algorithm \cite{neyshabur2015path} in terms of both accuracy and stability of the algorithm. To that end, we considered image classification problem with a path norm constraint on the network: $\|W_t\|_p \leq \lambda$ for varying $\lambda$ as before. We train a simple feed-forward network which consists of $2$ fully-connected hidden layers with $4000$ units each, followed by the output layer with $10$ nodes. We used ReLu nonlinearity as the activation function and cross entropy as the loss, see \cite{neyshabur2015path} for more details. \\ 
We performed experiments on 4 standard datasets for image classification: MNIST \cite{lecun1998gradient}, CIFAR (10,100) \cite{krizhevsky2012imagenet} and finally color images of house numbers from SVHN dataset \cite{netzer2011reading}.
Figure \ref{fig:pathcg} shows the result for  $\lambda=10^6$ (after tuning), it can achieve the same accuracy as that of Path-SGD. 

{\bf Path-CG has one main advantage over Path-SGD}:  our results  in the supplement that Path-CG is more stable while the path norm of Path-SGD algorithm increases rapidly. This shows that  Path-SGD \emph{does not} effectively regularize the path norm whereas Path-CG keeps the path norm less than $\lambda$ as expected.

\emph{{\bf Takeaway}: All statistical benefits of path norm are possible via CG and at the same time computationally more stable.}
\subsection{Image Inpainting using Conditional Gradients}
Finally, we illustrate the ability of our CG framework on an exciting and recent application of image inpainting using  Generative Adversarial Networks (GANs). We now briefly explain the overall experimental setup.  GANs using game theoretic notions can be defined as a system of 2 neural networks called Generator and the Discriminator competing with each other in a zero-sum game \cite{arora2017generalization}.  

Image inpainting/completion can be performed using the following two steps \cite{amos2016image}:  (i) Train a standard GAN as a normal image generation task, and (ii) use  the trained generator and then tune the noise that gives the best output, see Figure  \ref{fig:inpaint_ill} (left). Hence, our basic hypothesis is that if the generator is trained well, then the follow-up task of image inpainting benefits automatically.

{\bf Train DC-GAN faster for better image inpainting:} We used the state of the art DC-GAN architecture in our experiments and we impose a Frobenius norm constraint on the parameters but \emph{only} on the Discriminator to avoid mode collapse issues and trained using the CG algorithm.   In order to verify the performance of the CG algorithm, we used 2 standard face image datasets from CelebA and LWF and conducted two experiments: trained on the CelebA dataset with LFW being the test dataset and vice-versa.  The results are shown in Figure \ref{fig:inpaint_ill} (right) after tuning $\lambda$.   We found that the generator generates very high quality images after being trained with LFW images in comparison to the original DC-GAN \emph{in just $10$ epochs} (reducing the computational {\bf cost by $50\%$}), figure \ref{fig:inpaint_ill} shows that the images completed using the CG trained DC-GAN look  realistic. Our results indicate that CG trained DC-GAN qualitatively performs as good or better than the standard DC-GAN. Experiments showing the stability of our model with varying $\lambda$'s is in the supplement.

\emph{{\bf Takeaway}: GANs can be trained faster with no change in accuracy.}

\section{Conclusions}\vspace{-4pt}
The main emphasis of our work is to 
provide evidence supporting three distinct 
but related threads: 
(i) global constraints are relevant in the context of training deep models in vision and 
machine learning; (ii) the lack of support for global constraints in existing libraries like Keras and 
Tensorflow \cite{abadi2016tensorflow} may be because of the complex interplay between 
constraints and SGD which we have shown can be side-stepped, to a great extent, using CG; and 
(iii) constraints can be easily incorporated with negligible to small changes to existing implementations. 
We provide a range of empirical results on \emph{three} different case studies to support our claims, 
and conjecture that a broad variety of other problems will immediately benefit by viewing them through the 
lens of conditional gradient algorithms. 
Our analysis and experiments suggest concrete ways in which one 
may realize performance improvements, in both generalization and runtime, 
by substituting in CG schemes in certain classes of deep learning models. Tensorflow code for all our experiments will be made available in Github.

\bibliographystyle{natbib}
\bibliography{deep_cg}

\section*{Appendix}
\section{Dealing with Category 2 Constraints more efficiently} Let $\|W\|_{s,p}:=\|\left(\| W_1\|, \|W_2 \|,..., \|W_l\|\right)\|_p$ be the $\ell_{s,p}$ norm of $W$ where $W_i$ be the weight matrix corresponding to edges between layer $i-1$ and $i$. Then $\ell_1/\ell_{\infty}$ norm is defined as $\|W\|_{1,\infty}:=\max_i|W_i|_1$. Then we need to solve the following at each iteration:\begin{align}
\{s_i\}\in \arg\min_{W_i} \sum_{i=1}^l \langle G_i,W_i\rangle \quad \text{s.t.}\quad \max_i|W_i|_1\leq \lambda.
\end{align}
Now it is easy to see that we set that the optimal solution can be computed as follows. For each (gradient) matrix $G_i$ we simply compute the element (or index) with the maximum magnitude. We then set the corresponding index to $\lambda$ if the sign of the gradient in that index is negative and vice-versa. This procedure updates $l$ elements in each iteration. We can similarly switch the roles of the max function, absolute function and define the $\ell_{\infty}/\ell_1$ norm which will lead to updating one layer at each iteration.
For more specific group norms see \cite{garber2015faster}. Specifically, for matrix induced group norms, lemma 9 in \cite{garber2015faster} shows that the corresponding linear optimization problem at each iteration can be solved efficiently.
\section{Proof of Lemma 4.1.}
\begin{lemma}
	An $\epsilon$-approximate CG step for the TV norm can be computed  in $O(1/\epsilon)$ time (independent of dimensions of $A\in \R^{n\times E}$).
	\begin{proof} 
		\cite{harchaoui2015conditional} showed that in order to solve the CG step  for TV norm we first solve \begin{align}
		\max_{\beta\in\R,f} \beta~\text{s.t.}~ Af=\beta g_t, f\in [0,1]^E,\label{tv_cg}
		\end{align}
		where $E$ is the number of edges of the network and then compute the Lagrange multipliers at the optimal solution of problem \eqref{tv_cg}. Instead of first solving \eqref{tv_cg}, we propose to solve the dual directly using continuous optimization techniques. Introducing dual variables $v$ for the equality constraints and $w$ for the box constraints, our dual linear program can be written as,\begin{align}
		\max_{w,v} 1^Tw \quad \text{s.t.} \quad -A^Tv+w\leq 0, ~g_t^Tv=-1,~w\leq 0\label{main_eq}.
		\end{align}
		Since $v$ is a free variable, problem \eqref{main_eq} is equivalent to,\begin{align}
		\max_{w,v} 1^Tw \quad \text{s.t.} \quad w\leq A^Tv, ~g_t^Tv=1,~w\leq 0.\label{eq_1}
		\end{align}
		Eliminating $w$, we can write the dual of \eqref{tv_cg} as, \begin{align}
		\min_{v\in\R^E}\sum_{i=1}^E \min\left(0,v^TA_{[i,:]}\right)~\text{s.t.}~g_t^Tv=1
		\end{align}
		where $A_{[i,:]}$ denotes the $i$-th row of $A$. Now it is easy to see that many continuous optimization methods are amenable which does \emph{not} require us to instantiate the matrix $A$.  Hence we can now use Theorem 1 in \cite{johnson2013accelerating} gives us the desired result. 
	\end{proof}
\end{lemma}
\section{Path-CG} In this section we provide details on the Path-CG updates in Section 5 of the main paper. 
\begin{figure}[!h]
	\centering
	\includegraphics[scale=0.7]{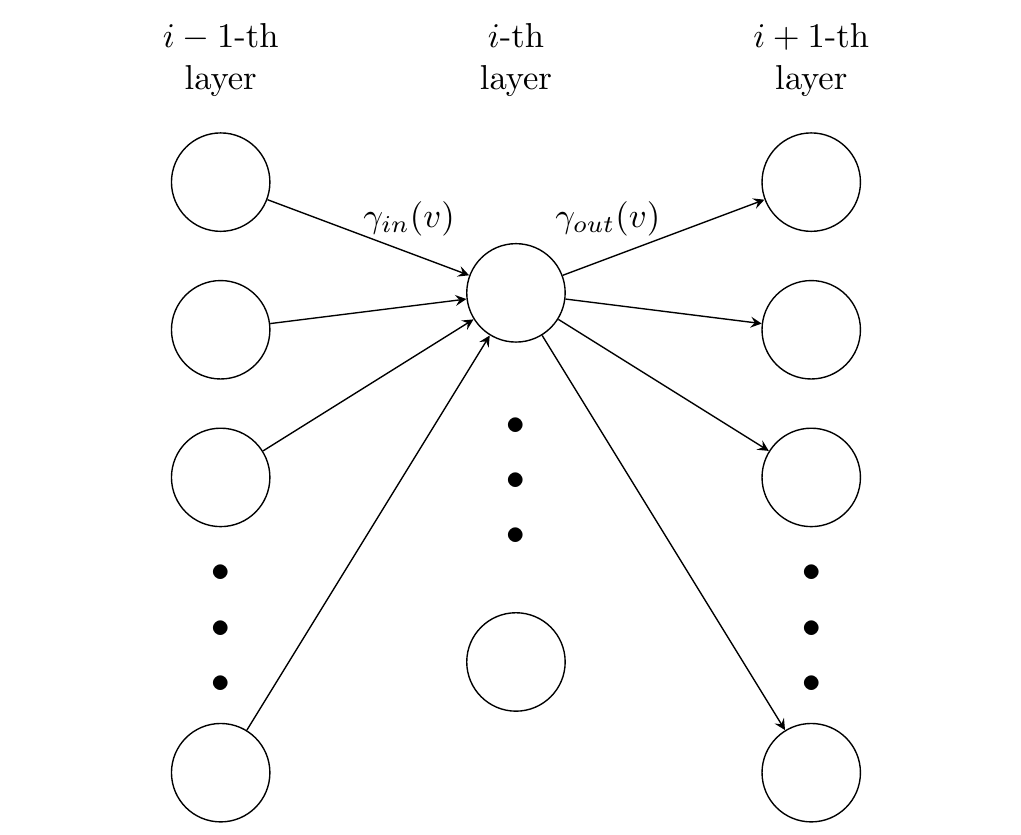}
	\caption{In order to compute the path norm, we can compute $\gamma_{in}$ and $\gamma_{out}$ for each node in the network. We then set $\gamma(u\to v)=\gamma_{in}(u)\gamma_{out}(v)$}
	\label{fig:tikz_path}
\end{figure}

Define $\gamma_{in}(v)$ and $\gamma_{out}(v)$ as, \begin{align}
\gamma_{in}(v) = \sum_{(u\to v)\in E}\gamma_{in}(u)w_{u,v}^2\\
\gamma_{out}(v) = \sum_{(v\to u)\in E}\gamma_{out}(u)w_{u,v}^2
\end{align}
Fix layer $i$ that we want to update and let the $g_i^t$ be the gradient of the loss function with respect to the weight matrix of the edges between $i-1$ and $i$ layers. Then the squared path norm  can be computed as the sum of squares (taken over all edges) of product of $\gamma_{uv}:=\gamma_{in}(u)\gamma_{out}(v)$ and $w_{uv}^2$   for every edge in between the layers, that is,\begin{align}
\|W\|_{\pi}^2:=\sum_{uv} \gamma_{uv}w_{uv}^2 = \|\hat{\Gamma}W\|_{F}^2
\end{align}
where the matrix $ \hat{\Gamma}$ is diagonal with elements being $\gamma_{uv}^{1/2}$. Hence we have that the Path-CG subproblems correspond to solving \begin{align}
\min_W g^TW \quad \text{s.t.}\quad \|\hat{\Gamma}W\|_2\leq \lambda
\end{align}
as shown in the main section.\\

\hrulefill

\end{document}